\documentclass{article}

\usepackage[utf8]{inputenc}
\usepackage[T1]{fontenc}
\usepackage{amsmath}
\usepackage{graphicx}
\usepackage{booktabs}
\usepackage{multirow}
\usepackage{natbib}
\usepackage{authblk}
\usepackage{url}

\DeclareMathOperator{\mean}{mean}
\DeclareMathOperator{\std}{std}

\title{Exploring the Potential of Symbolic Regression in Precision Livestock Farming}

\author{%
  E.~Mayrhuber$^{1,2}$, M.~Oczak$^{3,4}$, M.~Iwersen$^{5}$, D.~Brunner$^{1,2}$, V.~Dorfer$^{1}$, S.~Winkler$^{1}$\\
  \small $^{1}$\textit{Bioinformatics Research Group, PLFDoc, University of Applied Sciences}\\
  \small \textit{Upper Austria, Hagenberg, Austria}\\
  \small $^{2}$\textit{Computer Vision Lab, TU Wien, Vienna, Austria}\\
  \small $^{3}$\textit{Precision Livestock Farming Hub, The University of Veterinary Medicine Vienna, Vienna, Austria}\\
\small $^{4}$\textit{Center for Animal Nutrition and Welfare, The University of Veterinary Medicine Vienna, Vienna, Austria}\\
\small $^{5}$\textit{Center for Veterinary Systems Transformation and Sustainability, Clinical Unit for Herd Health Management, The University of Veterinary Medicine Vienna, Vienna, Austria}\\
    \small elisabeth.mayrhuber@fh-hagenberg.at%
}

\date{}

\begin{document}

\maketitle

\begin{abstract}
The ability to predict farrowing time in pigs is crucial for optimal farm management and ensuring piglet survival. This study investigates the potential of explainable artificial intelligence for precision livestock farming, focusing on acceleration sensor data from sows around the time of farrowing. Accurately predicting farrowing allows farmers to balance both, piglet safety by avoiding piglet crushing events and the mother sow's welfare, by avoiding early crating. Traditional machine learning models often lack interpretability (``closed-box'' problem). Explainable artificial intelligence offers a solution by revealing the decision-making processes within the models. This work is split into two stages. First the detection of nest building behaviour and second the prediction of the remaining time to farrowing. The nest building detection was possible for 166 out of 180 sows and successful within 48 hours before farrowing for 149 sows. The nest building detection is purely based on the change of a sow's activity level and uses no machine learning techniques. Further, this study utilizes symbolic regression, an open-box modelling technique that generates interpretable mathematical formulas, to predict the remaining time until the onset of farrowing. The results are compared with closed-box models, like support vector regression and artificial neural networks, for both explainability and performance. Notably, symbolic regression achieved a mean absolute error of 9.28 hours for the time-to-farrowing prediction, outperforming the closed-box models in terms of performance and interpretability. This research demonstrates that explainable artificial intelligence, specifically symbolic regression, can offer both interpretable results and competitive performance in precision livestock farming.

\end{abstract}

\section{Introduction}

Sows are routinely confined in crates a few days before giving birth (farrowing) mainly to prevent piglets from being crushed. While this practice protects piglets \citep{king2019}, it comes at the cost of increased stress and restricted natural behaviours for the mother sow \citep{hansen2017,heidinger2022,jarvis2006}. To balance piglet safety and sow wellbeing, modern farms have started to use temporary crating, delaying confinement previous to nest building and ideally at least 24 hours before farrowing \citep{cheon2022}. However, pinpointing the optimal time for crating to maximize welfare for both sows and piglets remains a challenge \citep{oczak2019}. This study proposes a two-step alarm system using explainable artificial intelligence (XAI) to address this challenge. The system will first detect the onset of nest building behaviour and then predict the time remaining until birth. We focus on symbolic regression (SR), an open-box machine learning (ML) technique within XAI. Unlike traditional ``closed-box'' models like artificial neural networks (ANNs) \citep{wang2003}, SR generates interpretable mathematical formulas, leveraging genetic programming \citep{affenzeller2009}. These formulas reveal the relationships between sensor data and the predicted time to farrowing, providing valuable insights for precision livestock farming (PLF) \citep{hoxhallari2022,najjar2023}. Genetic Programming has already been tested in other areas of PLF, such as beef cattle management \citep{abbona2020}. Building upon previous farrowing prediction research \citep{oczak2019}, this work aims to improve prediction accuracy by leveraging XAI with data from sow ear tag accelerometers. This approach differentiates itself from recent studies that rely on computer vision (CV) for farrowing prediction \citep{chen2023,oczak2023}.

\section{Material and Methods}

\subsection{Experimental data}

The study used data from 180 sows equipped with SMARTBOW\textsuperscript{\textregistered} ear tags\footnote{\url{www.smartbow.com}} (Smartbow/Zoetis LLC, Weibern, Austria). These tags contain tri-axial accelerometers that capture sow movement along three axes (x, y, and z). Details about the SMARTBOW\textsuperscript{\textregistered} system can be found in \citep{oczak2016}. As dataset, the magnitude of acceleration, calculated as the square root of the sum of squared values from each axis ($x^2 + y^2 + z^2$), was used. This single value represents the overall movement intensity of the sow. The dataset was split into training and test sets with 144 sows (80\%) and 36 sows (20\%), respectively (Table~\ref{tab:split}). This ensured an even distribution of data from different pen types (SWAP, trapezoid, and wing) across the test set. Evaluating the model's performance on the unseen test set allows us to assess its generalizability to new data.

\begin{table}
\centering
\caption{Number of sows used training and test split based on pen types.}\label{tab:split}
\begin{tabular}{lcc}
\toprule
\textbf{Pen type} & \textbf{Training} & \textbf{Test} \\
\midrule
Wing      & 48 & 12 \\
SWAP      & 44 & 12 \\
Trapezoid & 52 & 12 \\
\midrule
\textbf{Total} & \textbf{144} & \textbf{36} \\
\bottomrule
\end{tabular}
\end{table}

\subsection{Data Preprocessing and Feature Engineering}

The data preprocessing is different for the two tasks tackled in this research. The nest building detection works with rolling averages over the signal known so far. The averages that were calculated for this step are the rolling average over 12 hours, 24 hours and 48 hours. The 12-hour rolling average aims to show the day and night cycles of the sows, while the 24-hour and 48-hour windows aim to identify a general activity level for each sow and reveal larger changes in their activity. For the time-to-farrowing prediction, six features were extracted. These features include the sow's average daytime and nighttime activity, as well as the average activity over 24 and 48 hours during periods of normal day-night cycles. These values were calculated using the first 48 hours of available data for each sow. Additionally, the average activity over the last 12 and 24 hours were calculated at each timestamp. To create the training set, five time points were established for each sow. These timestamps were: three days after introduction to the farrowing pen, the timestamp of the nest building alarm, the timestamp between the start of nest building and the start of farrowing, one hour before farrowing, and the start of farrowing. The remaining minutes until farrowing were then calculated as target variable.

\subsection{Detection of Nest Building Behaviour}

For both the nest building detection and the time-to-farrowing prediction, the arrival of new acceleration data was simulated with a sliding window approach to test the models in an online setting. For every sliding window step the last 12 hours and the last day were compared to the signal which was received so far. The detection is only performed 2.5 days after the sow was introduced to the farrowing pen, since shortly after the introduction the activity levels can differ due to the change of environment \citep{oczak2019}. If one of the two conditions below (Equation~\ref{eq:cond1} and~\ref{eq:cond2}) evaluate to ``TRUE'', the detection of nest building behaviour alarm is triggered. The variable $a_t$ describes the acceleration at the time $t$ and $a_{t-24}$ is the acceleration 12 hours before the timestamp $t$. The constant values of 0.8 and 1.75 were chosen by trying out different combination for the training set and these achieved the best results.
\begin{align}
\mean\bigl([a_{t-24};\, a_t]\bigr) &\geq \Bigl(\mean\bigl([a_{t_0};\, a_{t-24}]\bigr) + \std\bigl([a_{t_0};\, a_{t-24}]\bigr)\Bigr) \cdot 0.8 \label{eq:cond1}\\
\max\bigl([a_{t-12};\, a_t]\bigr) &> \max\bigl([a_{t_0};\, a_{t-12}]\bigr) \cdot 1.75 \label{eq:cond2}
\end{align}

\subsection{Time to Farrowing Prediction}

Following the initial alarm trigger, time-to-farrowing predictions were performed every 90 minutes. Various machine learning algorithms were employed to predict the remaining time to farrowing. Two open-box algorithms, SR and linear regression (LR) were compared to four closed-box algorithms: ANNs, random forest, XGBoost (XGB), and support vector regression (SVR). For closed-box models, post-hoc explainability methods would be necessary to extract information regarding the reasoning behind model predictions or the features with a high impact on the results. Examples of such methods include LIME \citep{ribeiro2016} and SHAP \citep{lundberg2017}. The performance of all chosen models will be discussed in the ``Results and Discussion'' section, while explainability will not be evaluated for closed-box models. To train the models, the training dataset, which is described above, was used. The training of all models was performed using a grouped 5-fold cross-validation to ensure robust training. Grouping the data for cross-validation is important in this case since features were extracted from five timestamps and the data from one sow should not be split during cross validation.

For every model the same folds were used to ensure comparable results. EC-KitY \citep{sipper2023} was utilized to train SR models, while scikit-learn \citep{pedregosa2011} was used for all other models.

\section{Results and Discussion}

To evaluate the model performances during training a combination of the mean absolute error (MAE) and the Pearson correlation coefficient (PCC) were used. These training and validation performances can be seen in Table~\ref{tab:cv}. During the training phase the models have only been evaluated on the aforementioned predefined timestamps, the performance in a real-time scenario can be found in Table~\ref{tab:online}.

\begin{table}
\centering
\caption{Training and validation performance for 5-fold cross-validation.}\label{tab:cv}
\begin{tabular}{l*{8}{c}}
\toprule
\multirow{3}{*}{\textbf{Model}} & \multicolumn{4}{c}{\textbf{MAE [h]}} & \multicolumn{4}{c}{\textbf{PCC}} \\
\cmidrule(lr){2-5}\cmidrule(lr){6-9}
 & \multicolumn{2}{c}{\textbf{Training}} & \multicolumn{2}{c}{\textbf{Validation}} & \multicolumn{2}{c}{\textbf{Training}} & \multicolumn{2}{c}{\textbf{Validation}} \\
\cmidrule(lr){2-3}\cmidrule(lr){4-5}\cmidrule(lr){6-7}\cmidrule(lr){8-9}
 & \textbf{Mean} & \textbf{Std} & \textbf{Mean} & \textbf{Std} & \textbf{Mean} & \textbf{Std} & \textbf{Mean} & \textbf{Std} \\
\midrule
\textbf{SR}  & 17.70 & 1.21 & 22.55 & 2.25 & 0.60 & 0.05  & 0.42 & 0.06 \\
\textbf{LR}  & 26.80 & 0.85 & 27.22 & 2.42 & 0.45 & 0.01  & 0.44 & 0.04 \\
\textbf{ANN} & 23.17 & 1.03 & 24.56 & 2.22 & 0.56 & 0.02  & 0.50 & 0.04 \\
\textbf{RF}  & 8.83  & 0.43 & 24.35 & 2.73 & 0.96 & 0.004 & 0.49 & 0.06 \\
\textbf{XGB} & 0.64  & 0.10 & 26.01 & 1.66 & 0.99 & 6E-5  & 0.42 & 0.06 \\
\textbf{SVR} & 24.98 & 0.69 & 25.01 & 2.74 & 0.44 & 0.01  & 0.45 & 0.03 \\
\bottomrule
\end{tabular}
\end{table}

To demonstrate the effectiveness of our approach, we showcase the nest-building detection and the time-to-farrowing detection on one sow from the test set. The evaluation of the time-to-farrowing prediction was conducted in an online setting on both the training and test sets, using the best performing model from the cross-validation.

\subsection{Detection of Nest Building Behaviour}

\begin{figure}
\centering
\includegraphics[width=0.85\textwidth]{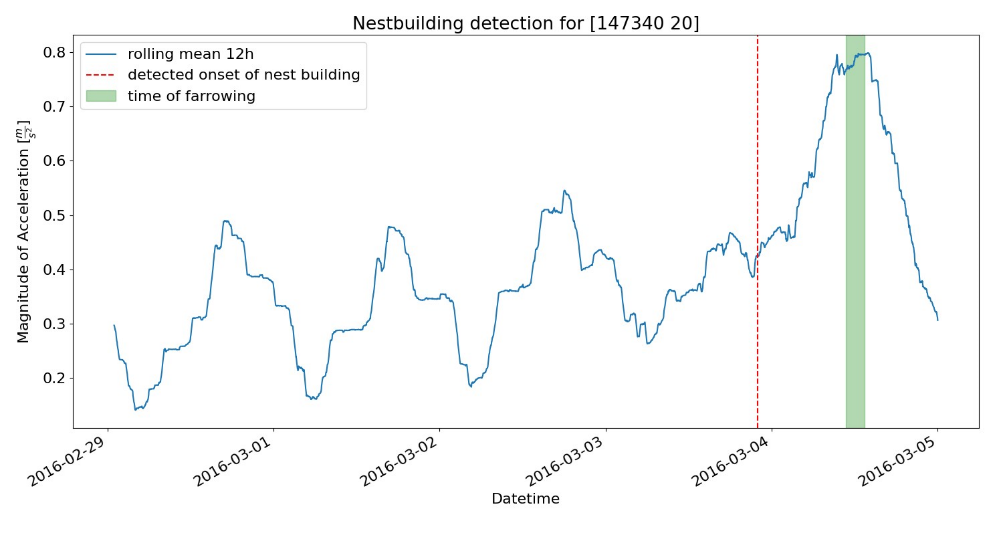}
\caption{Detection of nest building behaviour for one sow. The dashed line (red) shows the detected start of nest building behaviour, and the highlighted area (green) shows the true time of farrowing. This detection is done online using a sliding window approach.}\label{fig:detection}
\end{figure}

The detection of nest building behaviour before farrowing was possible for 166 of the 180 sows in our experiments. There was no case where the detection was after the onset of farrowing, but for 14 sows, no detection could be performed. In every successful case the alarm was triggered before the start of farrowing. In 17 cases the alarm was presented at least 48h before the start of farrowing and therefore more time to farrowing predictions were needed. In most cases this was caused by an anomaly in the sow's activity levels during normal day night cycles. Figure~\ref{fig:detection} shows the detection of nest building behaviour for the demonstration sow. This detection is robust to changes in the baseline activity levels of sows since it is only dependent on one sows' activity.

\subsection{Time to Farrowing Prediction}

For every time to farrowing prediction in the test setting, the same six features as for the training set were extracted from the signal that is known so far from each sow. This was then performed every 90 minutes to update the prediction, as new data arrives. This timespan can be easily adapted in the prototype, for example to every 60 minutes instead. Table~\ref{tab:online} shows quality metrics for every used model. It shows the average PCC and MAE, as well as the standard deviation for these metrics among the training and test set within this online prediction. The number of predictions differ for every sow, since the time between the detected onset of nest building and the actual time of farrowing varies a lot (6--188 hours).

Focussing on the MAE of the test sows (Table~\ref{tab:online}), SR achieves the best results, followed by SVR. A Mann-Whitney U test comparing the MAE of SR against SVR shows with a p-value of $2.18 \times 10^{-6}$ that there is a significant difference between these two. This means that SR, while being inherently more interpretable, also performs significantly better than SVR.

\begin{table}
\centering
\caption{Performance of different models in a real time setting for time to farrowing predictions.}\label{tab:online}
\begin{tabular}{l*{8}{c}}
\toprule
\multirow{3}{*}{\textbf{Method}} & \multicolumn{4}{c}{\textbf{MAE [h]}} & \multicolumn{4}{c}{\textbf{PCC}} \\
\cmidrule(lr){2-5}\cmidrule(lr){6-9}
 & \multicolumn{2}{c}{\textbf{Training}} & \multicolumn{2}{c}{\textbf{Test}} & \multicolumn{2}{c}{\textbf{Training}} & \multicolumn{2}{c}{\textbf{Test}} \\
\cmidrule(lr){2-3}\cmidrule(lr){4-5}\cmidrule(lr){6-7}\cmidrule(lr){8-9}
 & \textbf{Mean} & \textbf{Std} & \textbf{Mean} & \textbf{Std} & \textbf{Mean} & \textbf{Std} & \textbf{Mean} & \textbf{Std} \\
\midrule
\textbf{SR}  & 19.57 & 26.61 & 9.28  & 13.09 & 0.73 & 0.31 & 0.76 & 0.21 \\
\textbf{LR}  & 22.05 & 20.59 & 18.04 & 10.51 & 0.89 & 0.22 & 0.90 & 0.29 \\
\textbf{ANN} & 21.46 & 25.33 & 17.28 & 14.36 & 0.81 & 0.25 & 0.86 & 0.22 \\
\textbf{RF}  & 20.51 & 22.40 & 19.10 & 16.88 & 0.74 & 0.33 & 0.79 & 0.23 \\
\textbf{XGB} & 21.28 & 28.02 & 16.35 & 18.64 & 0.65 & 0.29 & 0.76 & 0.21 \\
\textbf{SVR} & 28.34 & 23.39 & 12.45 & 9.22  & 0.88 & 0.27 & 0.76 & 0.24 \\
\bottomrule
\end{tabular}
\end{table}

\begin{figure}
\centering
\includegraphics[width=0.45\textwidth]{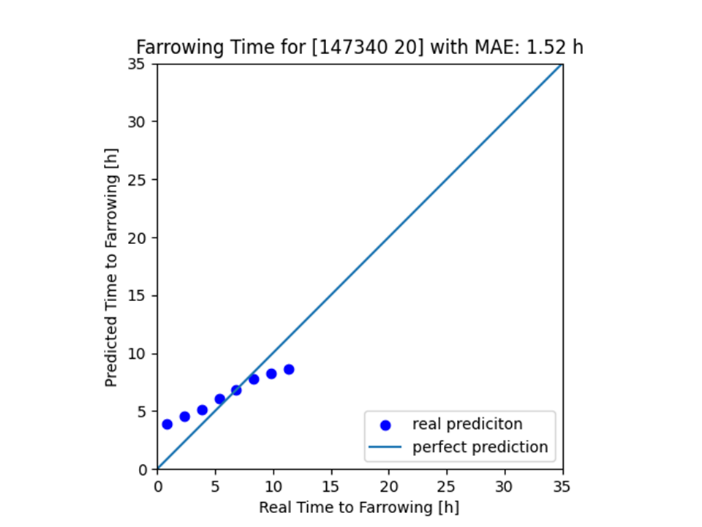}
\caption{The points show the predicted compared to the actual time to farrowing for one sow, using symbolic regression. The line shows how a perfect prediction would look like.}\label{fig:sr}
\end{figure}

Figure~\ref{fig:sr} shows every prediction that was performed using the SR model to predict the remaining time to farrowing. The actual prediction is done in minutes but for improved readability the values are transformed to hours. It shows that for the demonstration sow, the MAE is 1.52 hours and the PCC for this sow is 0.99, which is better than the average result.

\begin{figure}
\centering
\includegraphics[width=\textwidth]{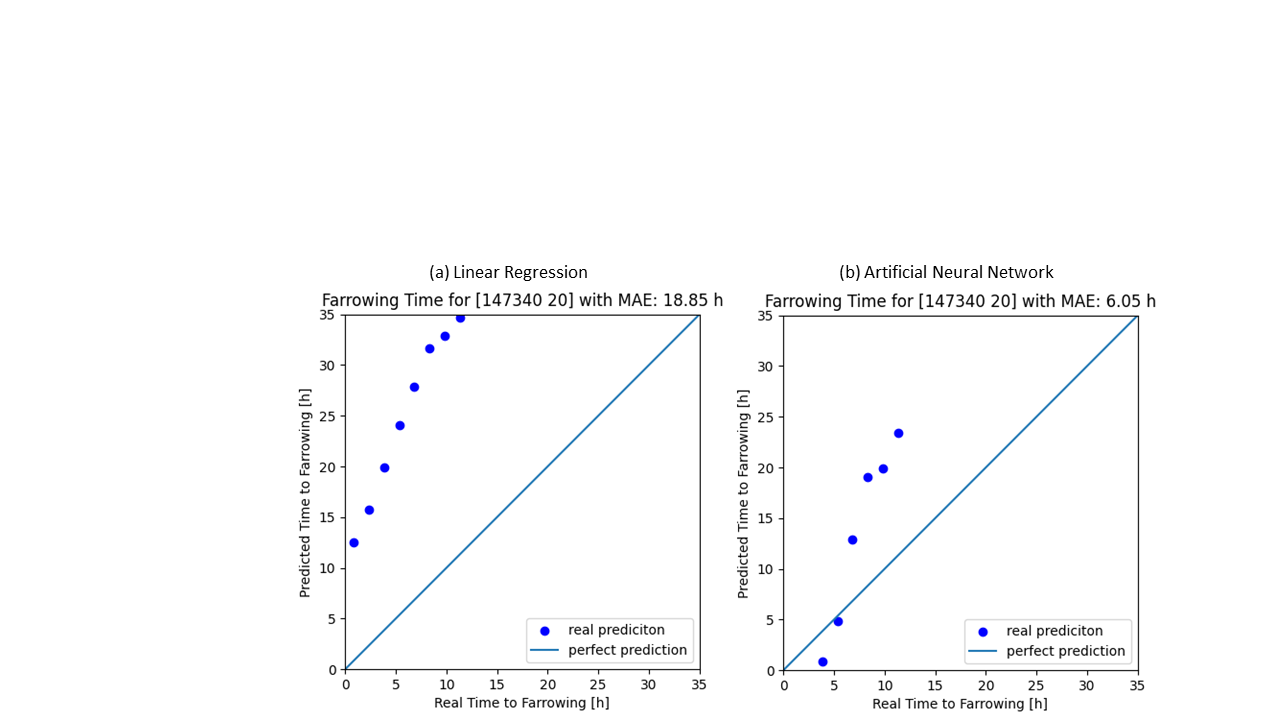}
\caption{Both images show the performance of the time to farrowing prediction for the same sow as in Figure~\ref{fig:sr}. The left figure (a) shows the prediction made by the linear regression model and the right figure (b) shows the prediction of the artificial neural network.}\label{fig:lrann}
\end{figure}

In contrast to Figure~\ref{fig:sr}, Figure~\ref{fig:lrann} shows the predictions for the same sow but with Linear Regression and the ANN model, which is a multi-layer perceptron neural network. It shows that the MAE for both models is worse than the one for the SR model. Also, the PCC is worse than that of the SR but this is not the case for every sow, which can be seen when having a look at Table~\ref{tab:online}, where the mean PCC for linear regression is higher than the one for SR.

\subsection{Explainability of Time to Farrowing Prediction}

In XAI literature there is often a distinction between explainability and interpretability \citep{marcinkevics2023}. Explainability means that a model's decision can be explained after it has been made, which is also called post-hoc explainability. On the other hand, interpretability means that the internals of a model can be interpreted and therefore decisions can be understood with no need of post-hoc explainability methods. Naturally SR builds interpretable models since the model is provided as an equation. The best resulting SR model of this research will not be provided in this paper, since the complexity of the model is too high to display efficiently. The used model has a total number of 2917 nodes when displayed as a tree. To show the structure of such models, Equation~\ref{eq:smallsr} shows a smaller SR model, where $f(t)$ is the remaining time to farrowing in minutes. The shown model has a significantly worse performance than the model which was used in the results above.
\begin{equation}
f(t) = 4683.5 + \left( \frac{0.467 \cdot \mean\bigl([a_{t-24};\, a_t]\bigr)}{-0.076 \cdot \mean\bigl([a_{t_0};\, a_{t-24}]\bigr)} \right) \cdot 306 \label{eq:smallsr}
\end{equation}

The two features included in this model are the mean magnitude of acceleration of the last 24 hours ($\mean([a_{t-24};\, a_t])$) and second, the average of the rolling mean over 24 hours except the last 24 hours ($\mean([a_{t_0};\, a_{t-24}])$).

\section{Conclusions}

For 166 out of 180 sows it was possible to provide an alarm at the beginning of nest building behaviour or shortly before. Compared to closed-box models like ANNs used in this study, SR offered a clear advantage in performance and interpretability. While achieving similar or even better performance in some cases, like it can be seen in the example above, ANNs lack the ability to interpret their decision-making process. This hinders user trust and limits the potential for knowledge discovery from the model. In conclusion the presented approach provides a robust, two stage, online farrowing prediction based on ear tag acceleration data. It consists of a first stage alarm at the beginning of nest building behaviour and from then on, an interpretable time to farrowing prediction, which is updated in a chosen interval. Since the models are based on the differences between the change of activities of individual sows the predictions are robust to changes in the baseline activity of sows, like it can be the case on different farms. While interpretable AI methods like SR provide valuable insights into a model's decisions, in its current form this mostly benefits ML experts. In future work, we aim to further enhance the interpretability of these methods for non-technicians like veterinarians.

\section*{Acknowledgements}

This research was funded in whole or in part by the Austrian Science Fund (FWF) [\url{https://doi.org/10.55776/DFH34}]. For open access purposes, the author has applied a CC BY public copyright license to any author-accepted manuscript version arising from this submission.

\bibliographystyle{plainnat}
\bibliography{references}

\end{document}